\documentclass[]{spie}  %>>> use for US letter paper
\usepackage{amsmath,amsfonts,amssymb}
\usepackage{graphicx}
\usepackage{subcaption}
\usepackage[colorlinks=true, allcolors=blue]{hyperref}

\title{Advancing Robotic Surgery: Affordable Kinesthetic and Tactile Feedback Solutions for Endotrainers}

\author[a]{Bharath Rajiv Nair}
\author[b]{Aravinthkumar T.}
\author[b]{B. Vinod}
\affil[a]{Columbia University, New York, USA}
\affil[b]{PSG College of Technology, Coimbatore, India}

\authorinfo{Further author information: (Send correspondence to Bharath Rajiv Nair)\\Bharath Rajiv Nair: E-mail: br2637@columbia.edu\\  Aravinthkumar T.: E-mail: aravinthkumar.thangavelu@gmail.com\\  B. Vinod: E-mail: hod.rae@psgtech.ac.in}

\begin{document} 
\maketitle

\begin{abstract}
The proliferation of robot-assisted minimally invasive surgery highlights the need for advanced training tools such as cost-effective robotic endotrainers. Current surgical robots often lack haptic feedback, which is crucial for providing surgeons with a real-time sense of touch. This absence can impact the surgeon's ability to perform delicate operations effectively.  To enhance surgical training and address this deficiency, we have integrated a cost-effective haptic feedback system into a robotic endotrainer. This system incorporates both kinesthetic (force) and tactile feedback, improving the fidelity of surgical simulations and enabling more precise control during operations. Our system incorporates an innovative, cost-effective Force/Torque sensor utilizing optoelectronic technology, specifically designed to accurately detect forces and moments exerted on surgical tools with a 95\% accuracy, providing essential kinesthetic feedback. Additionally, we implemented a tactile feedback mechanism that informs the surgeon of the gripping forces between the tool's tip and the tissue. This dual feedback system enhances the fidelity of training simulations and the execution of robotic surgeries, promoting broader adoption and safer practices.
\end{abstract}

\keywords{haptic feedback, kinesthetic feedback, tactile feedback, Force/Torque (F/T) sensor, surgical robot, robotic endotrainer, teleoperation}

\section{INTRODUCTION}
\label{sec:intro}  % \label{} allows reference to this section

Surgical robots are increasingly utilized for minimally invasive surgical procedures, allowing surgeons to operate through small incisions using specialized instruments and an endoscopic camera. These robots offer enhanced dexterity, greater precision, and reliability, while also minimizing the impact of hand tremors caused by surgeon fatigue. The benefits of robot-assisted surgery include reduced pain, fewer complications, quicker recovery times, and minimal scarring.

During manual minimally invasive surgery, surgeons use all their senses including their sense of touch to operate surgical tools. But, most of the surgical robots available today are teleoperated and have physically disconnected master side and slave side. This in turn does not provide any haptic feedback to the surgeon. Surgeons control the master side of the robot to perform the surgery relying completely on visual feedback from 3D cameras \cite{1_dion1997visual}. Although visual feedback provides helpful information for performing robotic surgery, the absence of haptic feedback is an obvious disadvantage in terms of safety and ease of usage\cite{3_bethea2004application, 2_van2009value}. 
 
Haptic feedback can be categorized into two types- kinesthetic (force) feedback and cutaneous (tactile) feedback \cite{4_hannaford2016haptics, 5_okamura2009haptic, 6_westebring2008haptics}. Kinesthetic feedback provides the surgeon a feel of the force acting on the surgical tool whereas tactile feedback provides a sense of the gripping forces and textures.
 
Many studies were conducted which led to the development of kinesthetic feedback systems for robot-assisted minimally invasive surgery. Implementing kinesthetic feedback requires precise measurement of the forces acting on the surgical tool at the patient’s side. The six-axis force/torque sensor developed by the German Aerospace Center (DLR) is probably one of the most advanced \cite{7_seibold2005prototype}. It utilises a Stewart platform for force measurements. Another popular and widely used Force/Torque sensor is the ATI Nano sensor (ATI Industrial Automation, Apex, NC, USA). These sensors have outstanding sensing capabilities. However, these Force/Torque sensors are very expensive. These sensors also come in predefined shapes and sizes, which makes it difficult to integrate the sensor for particular applications.  Hence, to overcome these limitations, a 3-axis Force/Torque sensor based on optoelectronic technology \cite{8_tar2011development, 9_noh2016multi} was developed. The light intensity based measurement using optoelectronic sensors has many advantages including high resolution, low power consumption, lesser noise and low cost. This sensor was designed keeping in mind the need to effectively integrate it with the surgical tool.

It is also extremely important for the surgeon to know the gripping force between the tool tip and the tissues \cite{5_okamura2009haptic}. Such information can be provided with the help of tactile feedback. Extensive research has been conducted on developing technologies for sensing forces \cite{new1_wade2017force, new3_eom2021embedded} and providing tactile feedback across various applications \cite{new2_piacenza2018data, new4_o2018stretchable}. Tactile sensors and tactile feedback systems have been designed to provide surgeons with information about the material characteristics of the tissues that the surgical tool is in contact with \cite{10_takashima2005endoscopic, 11_lim2015role}. Hence, a simple tactile feedback system was designed which provides vibration feedback that varies with gripping force between the surgical tool tip and tissues. Incorporation of tactile feedback in the surgical robot will reduce tissue damage and breakage and hence will reduce healing time of the patient.

\section{THE ROBOTIC ENDOTRAINER}
The robotic endotrainer (or surgical robot training system) was developed with the aim of providing actual robot based surgical training for surgeons at a low cost. Hence, the design is very similar to that of the daVinci Si surgical robot (Intuitive Surgical, Mountain View, CA, USA).

\subsection{Master Side}
\begin{figure}[ht]
    \centering
    \begin{subfigure}{0.45\textwidth} % Adjust the width as needed
        \centering
        \includegraphics[height=6cm]{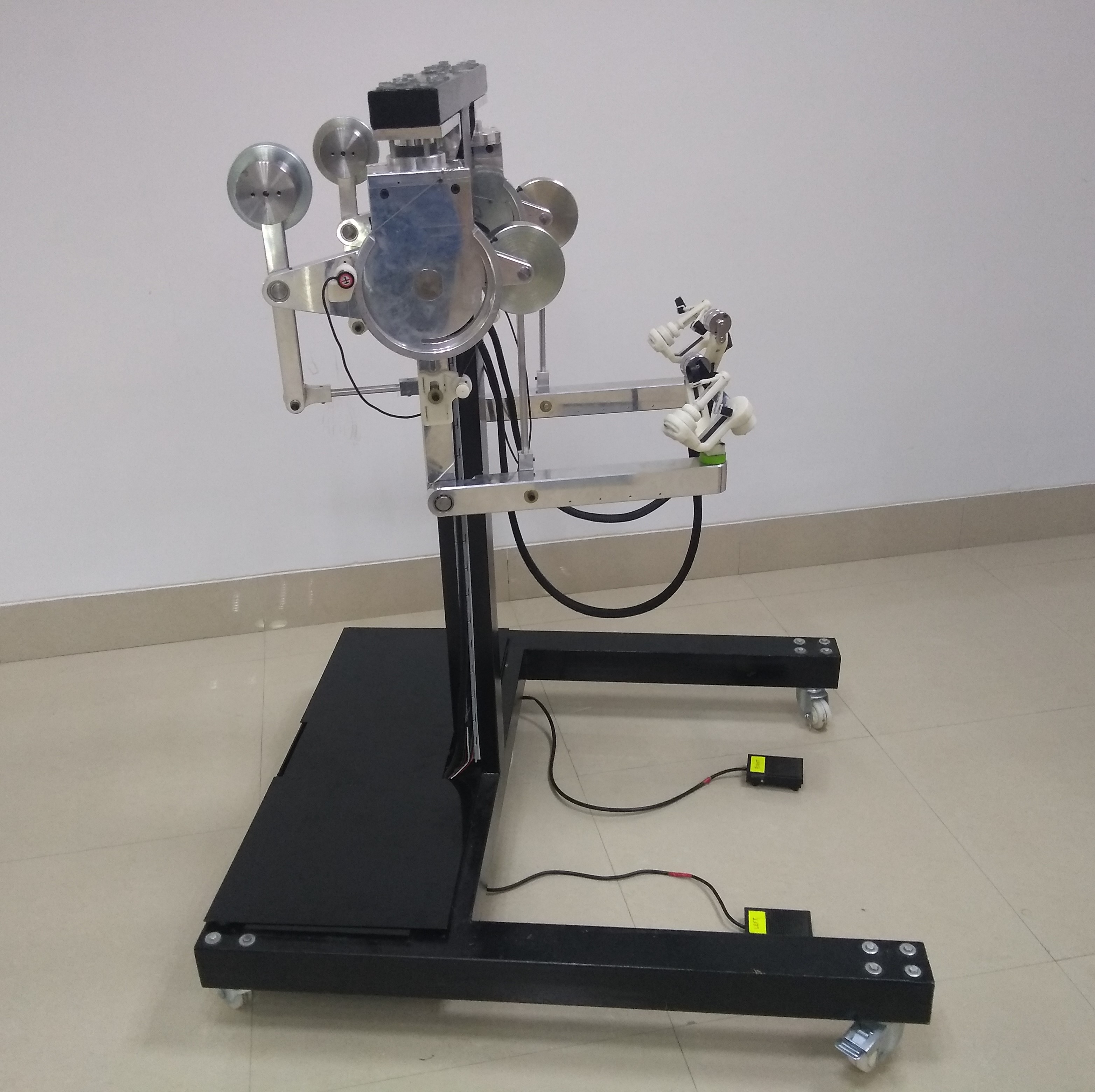}
        \caption{Prototype}
        \label{fig:master1}
    \end{subfigure}
    \hfill % Add some space between the subfigures
    \begin{subfigure}{0.45\textwidth} % Adjust the width as needed
        \centering
        \includegraphics[height=6cm]{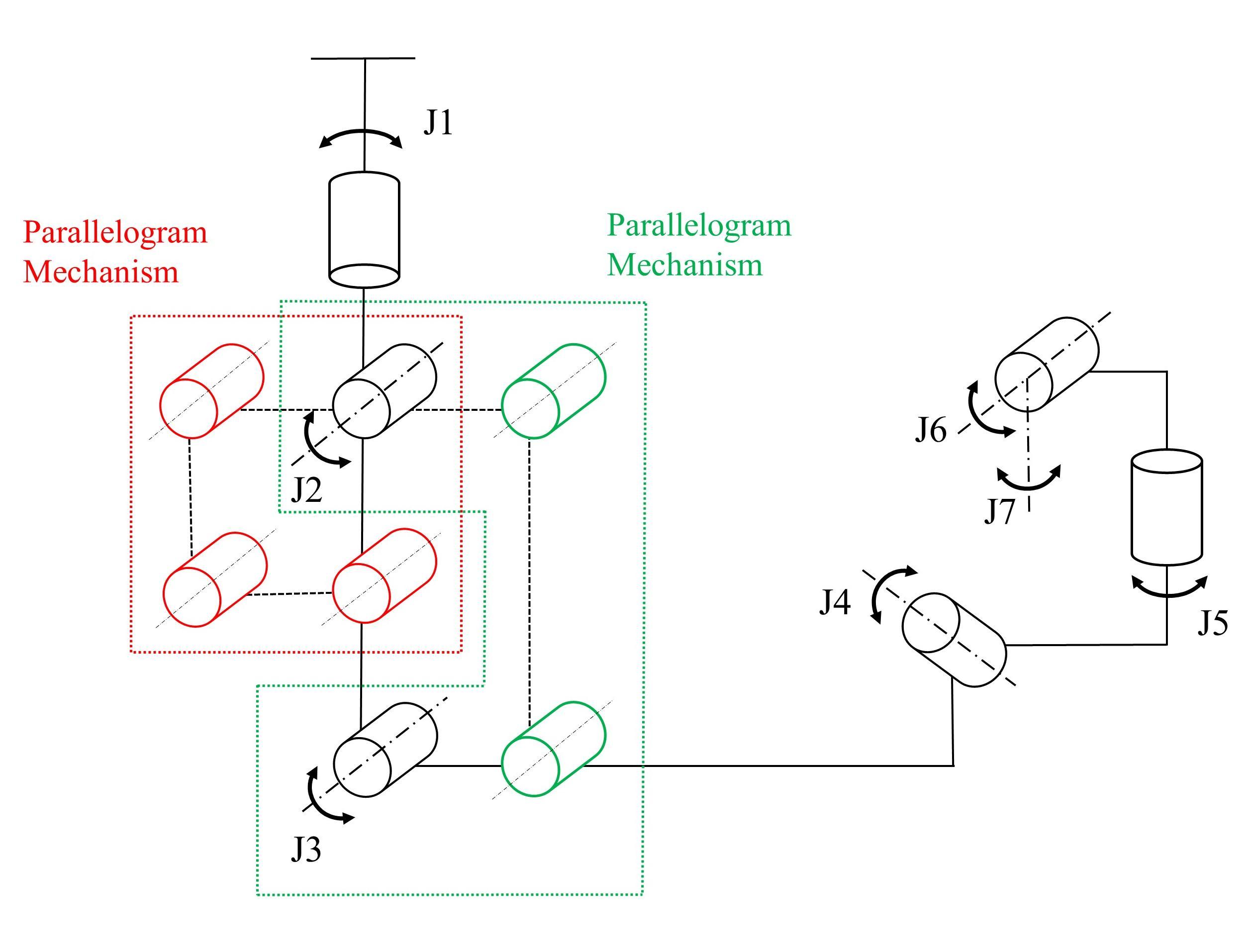} % Replace "your_second_image.eps" with the filename of your second image
        \caption{Structural Diagram}
        \label{fig:master2}
    \end{subfigure}
    \caption{Master side of the robotic endotrainer.}
    \label{fig:master_endo}
\end{figure}

The master side of the robotic endotrainer has two arms to control two manipulators on the slave side. The human arm has seven degrees of freedom (DOFs), hence each of the master arms needs at least seven DOFs. The master arms for the robot have seven DOFs, three for positioning and four for orientation. The prototype and structural diagram of the master side of the robotic endotrainer is shown in Fig. \ref{fig:master_endo}.

The joints utilized for positioning are equipped with incremental rotary encoders (Autonics Corporation, South Korea) and the joints utilized for orientation are equipped with magnetic rotary encoders (Renishaw PLC, United Kingdom) to measure the joint angles. Commercially available gimbal
motors are attached to the master manipulator joints to provide haptic feedback based on the force sensed by the novel force/torque sensor on the surgical tool.

\subsection{Slave Side}
The slave side includes two slave manipulators for handling two different surgical instruments, along with one camera arm for controlling an endoscope camera. Each slave arm features seven DOFs, with three dedicated to positioning and four for orientation.

\begin{figure}[ht]
    \centering
    \begin{subfigure}{0.45\textwidth} % Adjust the width as needed
        \centering
        \includegraphics[height=6cm]{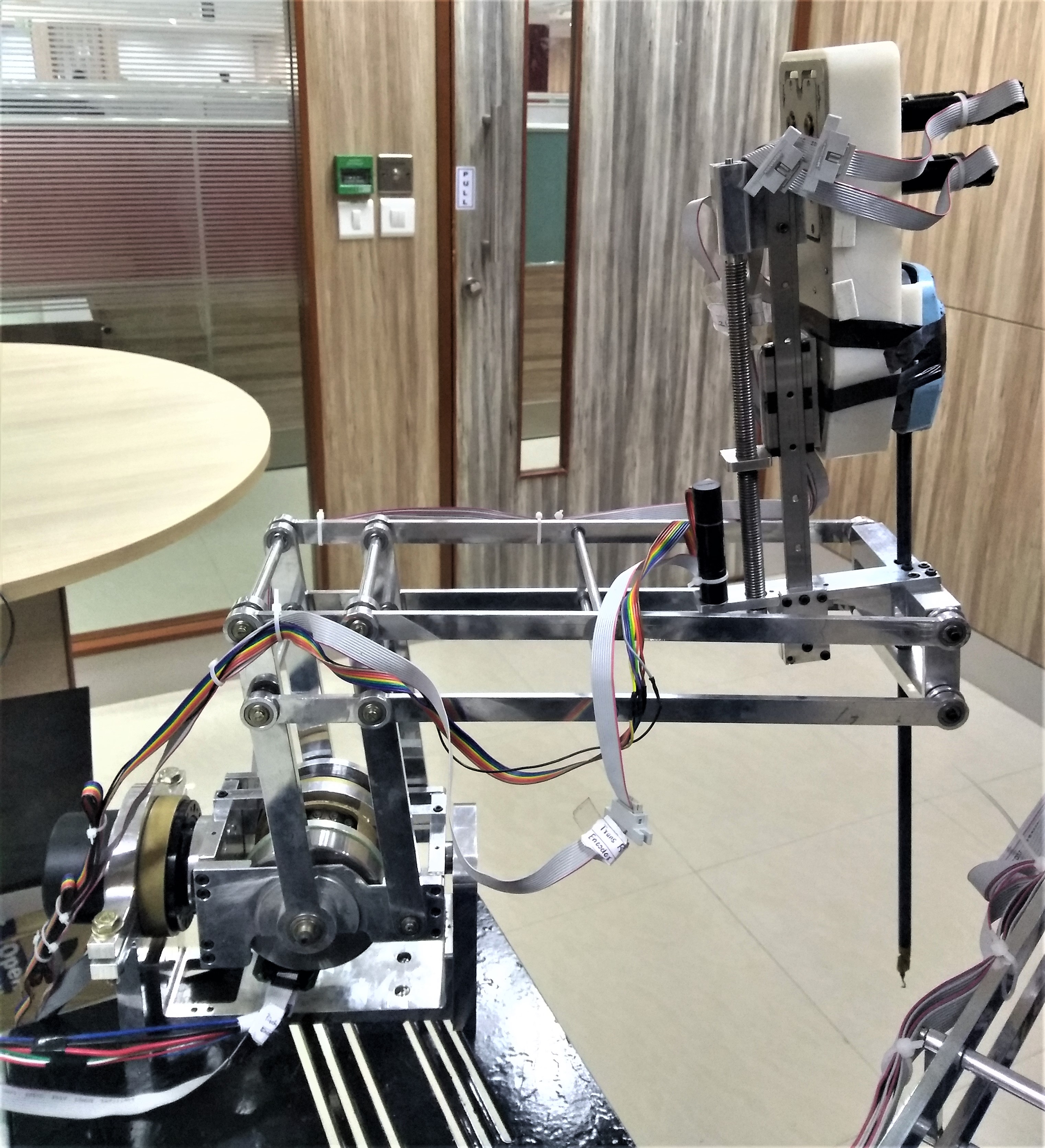}
        \caption{Prototype}
        \label{fig:master1}
    \end{subfigure}
    \hfill % Add some space between the subfigures
    \begin{subfigure}{0.45\textwidth} % Adjust the width as needed
        \centering
        \includegraphics[height=6cm]{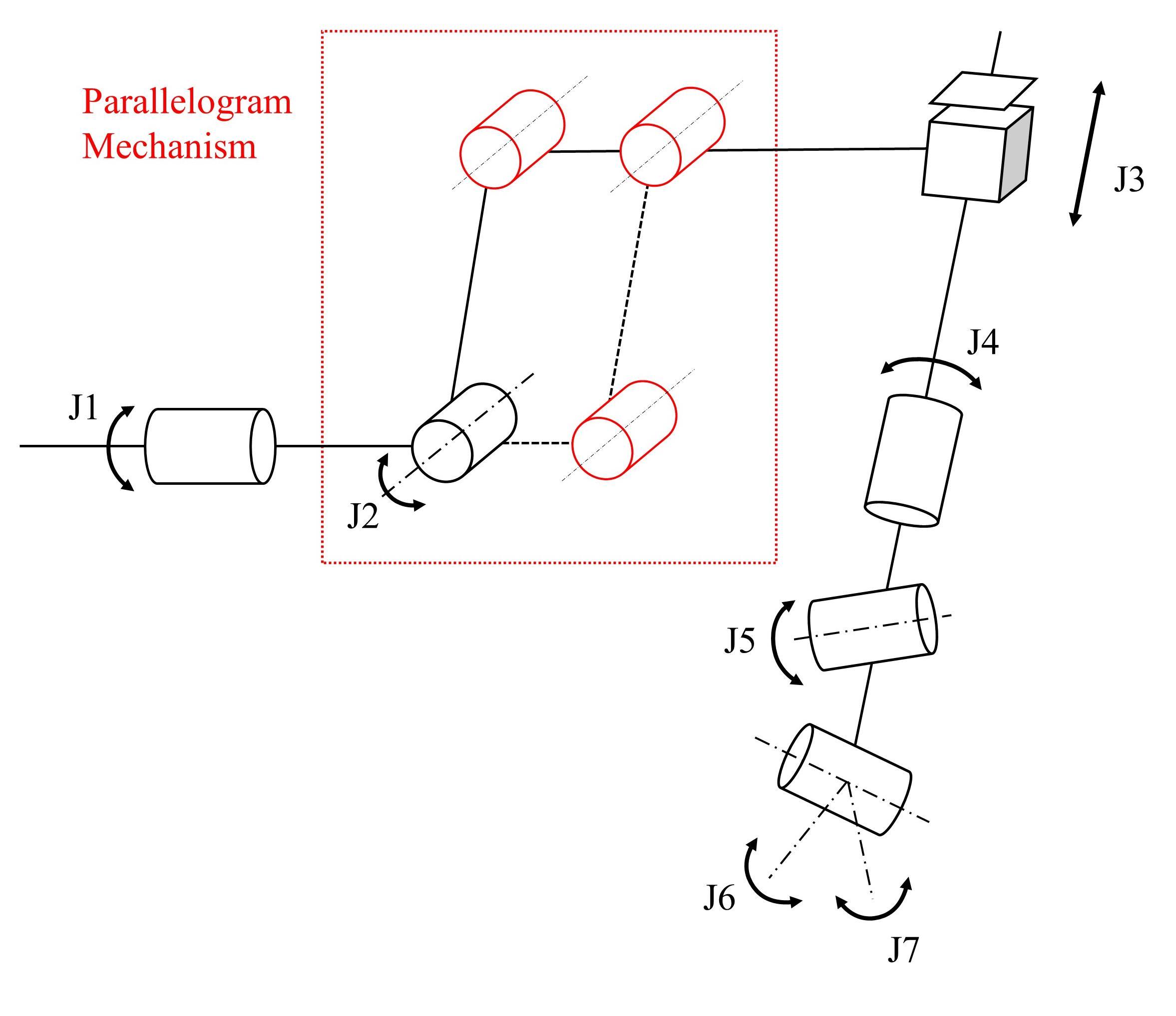} % Replace "your_second_image.eps" with the filename of your second image
        \caption{Structural Diagram}
        \label{fig:master2}
    \end{subfigure}
    \caption{Slave side of the robotic endotrainer.}
    \label{fig:slave_endo}
\end{figure}

All the joints in slave manipulators are equipped with DC
motors (Maxon, Switzerland) for precise control. Remote
Centre of Motion \cite{12_aksungur2015remote} is achieved with the help of
parallelogram mechanism. The camera arm has four DOFs,
three for positioning of the endoscope camera and one for
orientation. The joints are equipped with DYNAMIXEL
actuators (Robotis, USA). The prototype and
structural diagram of the slave side of the robotic endotrainer is shown in Fig. \ref{fig:slave_endo}.

\subsection{System Architecture}

\begin{figure} [hb]
   \begin{center}
   \begin{tabular}{c} %% tabular useful for creating an array of images 
   \includegraphics[height=8cm]{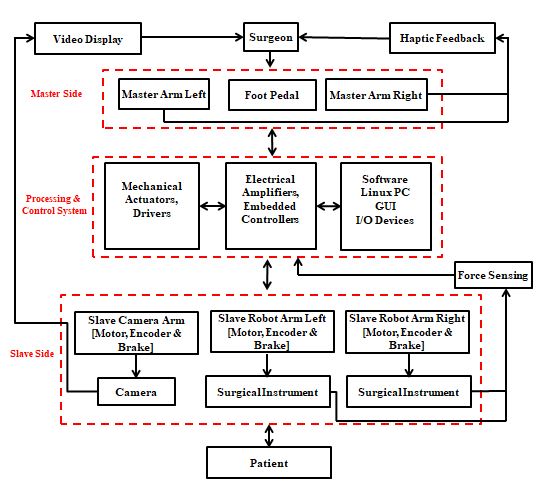}
   \end{tabular}
   \end{center}
   \caption[example] 
   { \label{fig:system_arch} 
System Architecture}
   \end{figure} 

The flowchart of system architecture is shown in Fig. \ref{fig:system_arch}. The surgeon's movements on the master side are recorded by
the encoders and sent to a Linux based PC. Based on these
inputs, the controller will enable a similar but more precise
and scaled down motion of the slave manipulators. The
surgeon can also switch between controlling the slave
manipulators and the endoscopic camera arm by using the foot
pedal.

The forces sensed by the force/torque sensor is sent to the
PC which computes the individual torques to be exerted by
each motor on the master side to provide kinesthetic feedback.
Similarly, the vibration intensity of the vibration motor on the
master side is computed based on the sensed gripping forces
between tool tip and tissues to provide tactile feedback.

\section{KINESTHETIC FEEDBACK}
The motors on the master arm should generate the same
force and torques as that experienced by the slave manipulator in order to provide kinesthetic feedback. A three axis force/torque sensor based on optoelectronic technology was developed to measure the force acting on the slave manipulator. The torque required to be applied by the motors at each joint of the master arm are then computed based on the sensed forces. The combined effect of the joint torques at the master side produces forces similar to that experienced by the slave manipulator.

\subsection{Mechanical Design of Force/Torque Sensor}

\begin{figure} [h]
   \begin{center}
   \begin{tabular}{c} %% tabular useful for creating an array of images 
   \includegraphics[height=7cm]{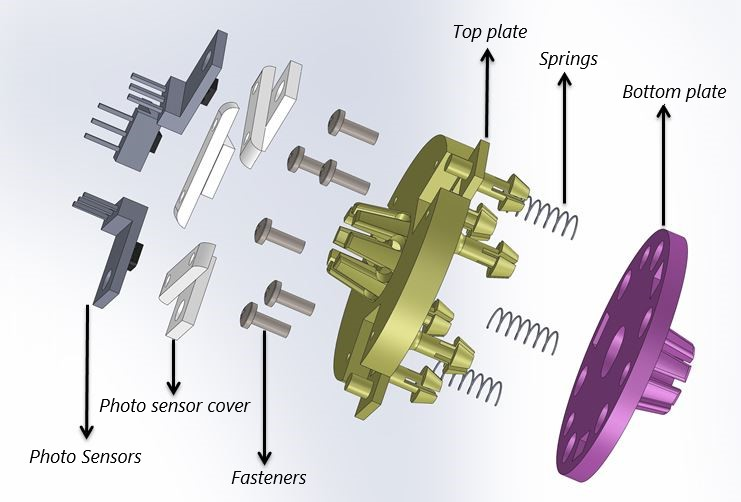}
   \end{tabular}
   \end{center}
   \caption[example] 
   { \label{fig:FTexplodedview} 
Exploded view of force/torque sensor. }
   \end{figure} 
   
A unique mechanical design was adopted for the three-axis
Force/Torque sensor to measure the forces and torques acting
on the slave side surgical tool. The exploded view
of the force/torque sensor can be seen in Fig. \ref{fig:FTexplodedview}. The main components of the
force/torque sensor are the top plate, bottom plate, three
springs and three photo sensors.

The developed sensor has an outer diameter of 40mm and a
height of 28mm. The sensor also has a through hole with a
diameter of 8.5mm to allow for the passage of all the strings
within the surgical tool. The upper and lower plates of force/torque sensor are made
up of PLA (Polylactic Acid) type of plastic. These
components were 3D printed by using the Fused Deposition
Modeling (FDM) technique, which is fast and cheap. Stainless steel springs are used to maintain a gap between
the sensor plates. The deflections of the springs measured by
the photo sensors are directly proportional to the magnitude of
force and torque acing on the instrument.

\begin{figure}[ht]
    \centering
    \begin{subfigure}{0.45\textwidth} % Adjust the width as needed
        \centering
        \includegraphics[height=5cm]{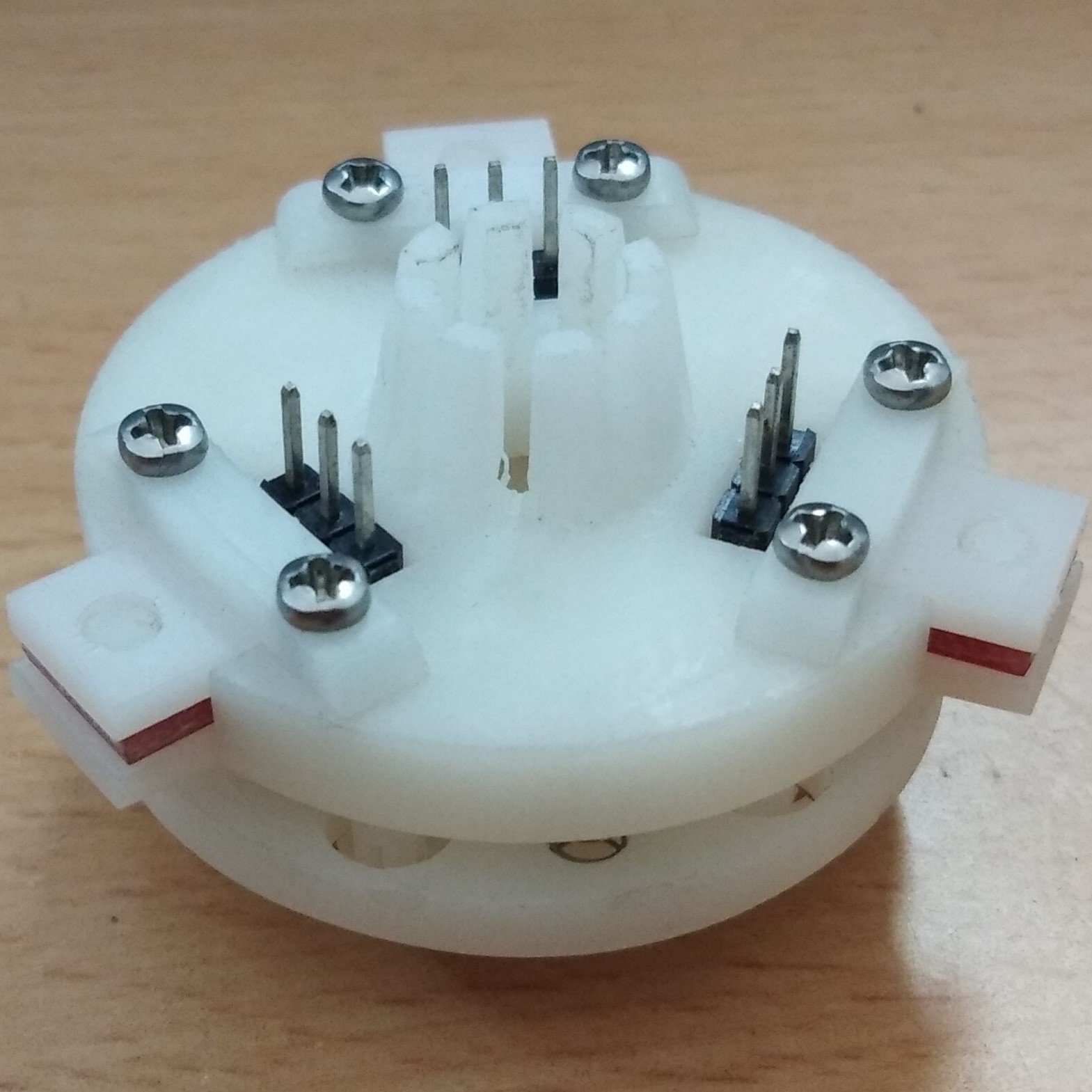}
        \caption{Isometric View}
        \label{fig:master1}
    \end{subfigure}
    \hfill % Add some space between the subfigures
    \begin{subfigure}{0.45\textwidth} % Adjust the width as needed
        \centering
        \includegraphics[height=5cm]{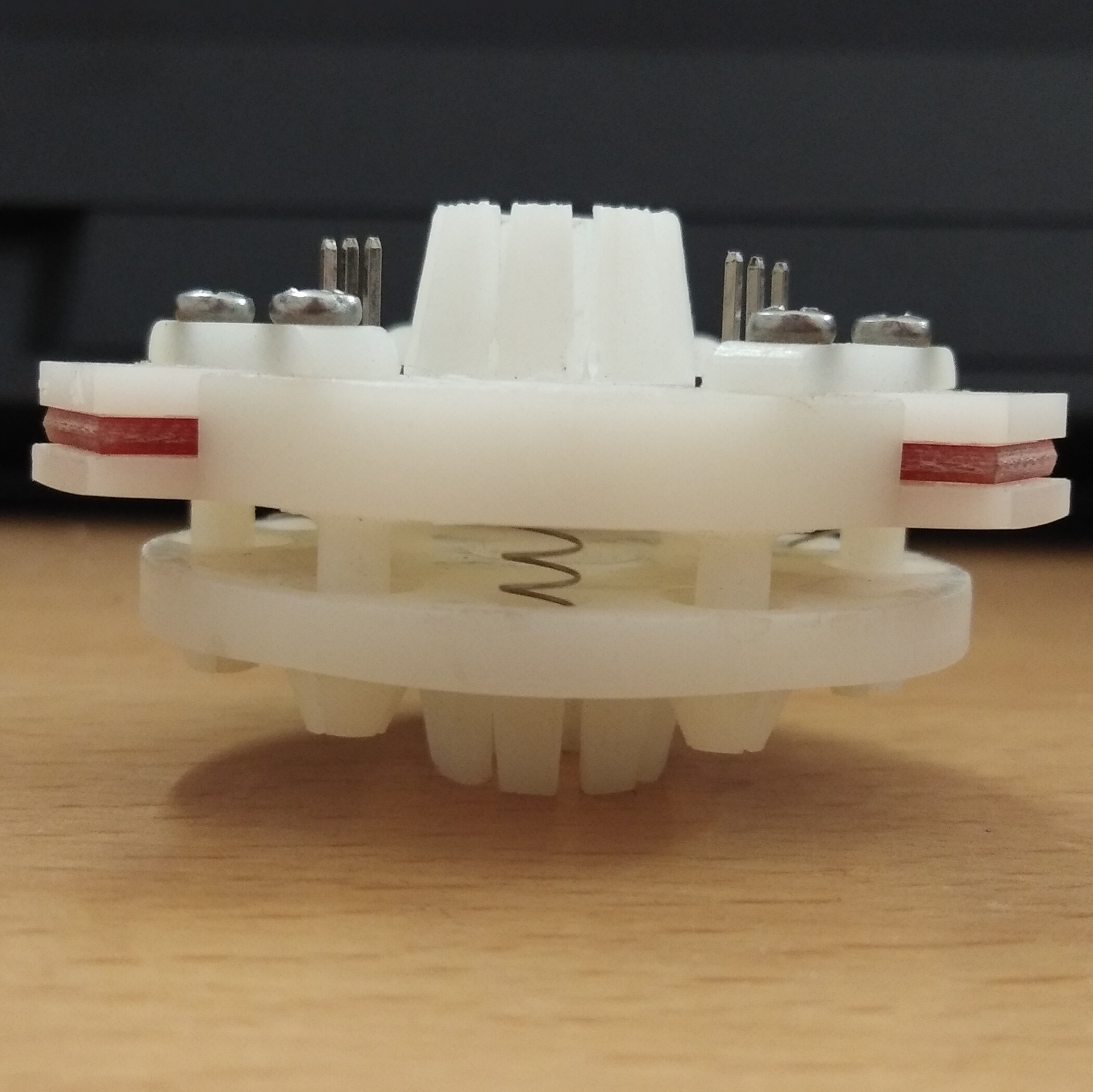} % Replace "your_second_image.eps" with the filename of your second image
        \caption{Front view}
        \label{fig:master2}
    \end{subfigure}
    \caption{3D printed F/T sensor.}
    \label{fig:FT_Views}
\end{figure}

A snap fit mechanism was adopted to attach the top plate
with the bottom plate. The snap fit mechanism allows for
relative motion of the top and bottom plates when a force is
applied. The snap fit mechanism also allows easy assembling
and dismantling of the sensor. The developed
three-axis force/torque sensor is shown in Fig. \ref{fig:FT_Views}.

\begin{figure} [h]
   \begin{center}
   \begin{tabular}{c} %% tabular useful for creating an array of images 
   \includegraphics[height=5cm]{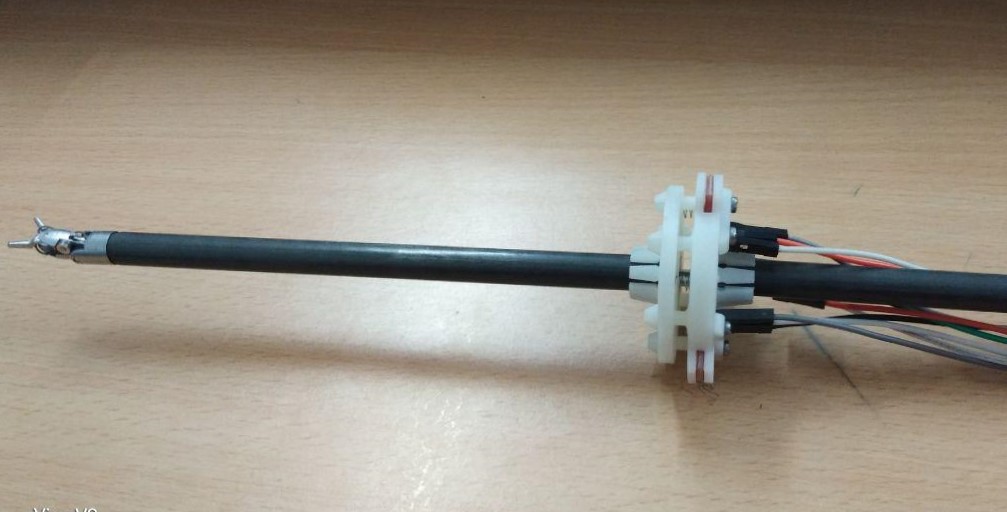}
   \end{tabular}
   \end{center}
   \caption[example] 
   { \label{fig:FT_tool} 
Force/Torque sensor integrated with surgical instrument. }
   \end{figure} 

The surgical instrument was cut horizontally to incorporate
the force/torque sensor for effective measurement of the force
and torque. The collet mechanism on each sensor plate holds
the outer casing of the surgical tool tightly without the use of
any fastener. The integration of the force/torque sensor with the surgical instrument is depicted in Fig. \ref{fig:FT_tool}.

\subsection{Sensing Forces and Torques}

\begin{figure}[ht]
    \centering
    \begin{subfigure}{0.45\textwidth} % Adjust the width as needed
        \centering
        \includegraphics[height=5cm]{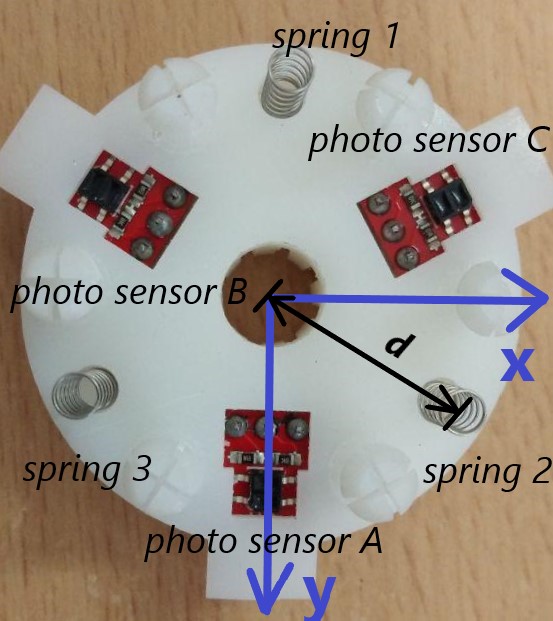}
        \caption{Top View of cross-section}
        \label{fig:master1}
    \end{subfigure}
    \hfill % Add some space between the subfigures
    \begin{subfigure}{0.45\textwidth} % Adjust the width as needed
        \centering
        \includegraphics[height=5cm]{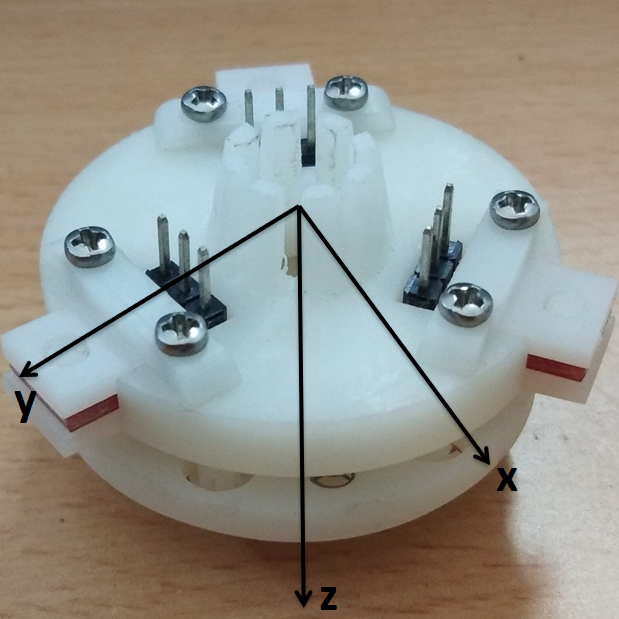} % Replace "your_second_image.eps" with the filename of your second image
        \caption{Isometric view}
        \label{fig:master2}
    \end{subfigure}
    \caption{Sensor structure, design variables and assumed axes.}
    \label{fig:axis_sensor}
\end{figure}

The developed sensor is capable of measuring the value of
force acting along the z-axis and moments along x-axis and y-axis.
The structure of the force/torque sensor can be simplified to
three springs symmetrically arranged within the sensor at an
angle of 120° between them. The three photo sensors are
placed diametrically opposite to the springs to measure the
deflection of the springs. The positioning of the springs, photo
sensors and assumed direction of the z-axis, x-axis and y-axis
is shown in Fig. \ref{fig:axis_sensor}. When an external force or moment is
applied, each of the three springs is deflected. The photo
sensors can measure the deflections of the three springs. The
force and moments are then computed from the values of
spring deflections.

Let $k$ represent the stiffness of each of the springs and $d$
represent the radial distance of the springs from the center of
the sensor.

In the presence of only $F_z$, all the springs will deflect by the
same amount. This deflection of the springs in the presence of
$F_z$ is denoted by $\delta_{1Fz}$, $\delta_{2Fz}$ and $\delta_{3Fz}$.

\begin{equation}
\label{eq:1}
F_z = 3k\delta
\end{equation}

\begin{equation}
\label{eq:2}
\delta_{1Fz} = \delta_{2Fz} = \delta_{3Fz} = \frac{F_z}{3k}
\end{equation}

In the presence of only $M_x$, $spring 1$ will elongate and $spring 2$ and $spring 3$ will compress. This deflection of the springs in the presence of $M_x$ is denoted by $\delta_{1Mx}$, $\delta_{2Mx}$ and $\delta_{3Mx}$.

\begin{equation}
\label{eq:3}
M_x = k.\delta_{1M_x}.d - k.\delta_{2M_x}.\frac{d}{2} - k.\delta_{3M_x}.\frac{d}{2}
\end{equation}

From basic trigonometry,

\begin{equation}
\label{eq:4}
\delta_{1M_x} = -2\delta_{2M_x} = -2\delta_{3M_x}
\end{equation}

\begin{equation}
\label{eq:5}
\delta_{1M_x} = \dfrac{2M_x}{3kd} ,   \qquad  \delta_{2M_x} = -\dfrac{M_x}{3kd} ,  \qquad  \delta_{3M_x} = -\dfrac{M_x}{3kd}
\end{equation}

In the presence of only $M_y$, $spring 2$ will elongate and
$spring 3$ will compress by equal amounts due to symmetry.
$spring 1$ will not undergo any deflection. The deflection of the springs in the presence of $M_y$ is denoted by $\delta_{1M_y}$, $\delta_{2M_y}$ and $\delta_{3M_y}$.

\begin{equation}
\label{eq:6}
M_y = k.\delta_{2M_y}.\dfrac{\sqrt{3}d}{2} - k.\delta_{3M_y}.\dfrac{\sqrt{3}d}{2}
\end{equation}

\begin{equation}
\label{eq:7}
\delta_{1M_y} = 0 ,   \qquad  \delta_{2M_y} = \dfrac{M_y}{\sqrt{3}kd} ,  \qquad  \delta_{3M_y} = -\dfrac{M_y}{\sqrt{3}kd}
\end{equation}

The total deflection of the three springs when acted upon by
a combination of $F_z$, $M_x$ and $M_y$ is represented by $\delta_1$, $\delta_2$ and $\delta_3$.

\begin{equation}
\label{eq:8}
\delta_1 = \delta_{1F_z} + \delta_{1M_x} + \delta_{1M_y} = \dfrac{1}{3k}F_z + \dfrac{2}{3kd}M_x
\end{equation}

\begin{equation}
\label{eq:9}
\delta_2 = \delta_{2F_z} + \delta_{2M_x} + \delta_{2M_y} = \dfrac{1}{3k}F_z - \dfrac{1}{3kd}M_x + \dfrac{1}{\sqrt{3}kd}M_y
\end{equation}

\begin{equation}
\label{eq:10}
\delta_3 = \delta_{3F_z} + \delta_{3M_x} + \delta_{3M_y} = \dfrac{1}{3k}F_z - \dfrac{1}{3kd}M_x - \dfrac{1}{\sqrt{3}kd}M_y
\end{equation}

Since the photo sensors are placed diametrically opposite to
the springs, the change in distance measured by the sensors
will be equal and opposite to the deflection of the springs.

\begin{equation}
\label{eq:11}
\delta_A = -\delta_1, \qquad \delta_B = -\delta_2, \qquad \delta_C = -\delta_3
\end{equation}

Representing equations \eqref{eq:8} , \eqref{eq:9}, \eqref{eq:10}, \eqref{eq:11} in matrix form, we get:

\begin{equation}
\label{eq:12}
\begin{bmatrix}
\delta_A \\
\delta_B \\
\delta_C
\end{bmatrix} 
= -
\begin{bmatrix}
\frac{1}{3k} & \frac{2}{3kd} & 0 \\
\frac{1}{3k} & -\frac{1}{3kd} & \frac{1}{\sqrt{3}kd} \\
\frac{1}{3k} & -\frac{1}{3kd} & -\frac{1}{\sqrt{3}kd}
\end{bmatrix}
\begin{bmatrix}
F_z \\
M_x \\
M_y
\end{bmatrix}
\end{equation}

The matrix equation that estimates the values of $F_z$, $M_x$ and
$M_y$ when given the values of $\delta_A$, $\delta_B$ and $\delta_C$ is shown below in equation \eqref{eq:13}. On substituting the values of the spring constant $k =
0.196N/mm$ and the distance between the center of the sensor
to the spring $d = 16mm$, we get:

\begin{equation}
\label{eq:13}
\begin{bmatrix}
F_z \\
M_x \\
M_y
\end{bmatrix}
= -
\begin{bmatrix}
\frac{1}{3k} & \frac{2}{3kd} & 0 \\
\frac{1}{3k} & -\frac{1}{3kd} & \frac{1}{\sqrt{3}kd} \\
\frac{1}{3k} & -\frac{1}{3kd} & -\frac{1}{\sqrt{3}kd}
\end{bmatrix}^{-1}
\begin{bmatrix}
\delta_A \\
\delta_B \\
\delta_C
\end{bmatrix} 
= -
\begin{bmatrix}
0.196 & 0.196 & 0.196 \\
3.135 & -1.567 & -1.567 \\
0 & 2.717 & -2.717
\end{bmatrix}
\begin{bmatrix}
\delta_A \\
\delta_B \\
\delta_C
\end{bmatrix}
\end{equation}

Here, the force is obtained in Newtons and the torques
are obtained in N-mm.

\subsection{Force and Torque Sensing: Experimental Findings}

\begin{figure}[h]
    \centering
    \begin{subfigure}{0.31\textwidth} % Adjust the width as needed
        \centering
        \includegraphics[height=4cm]{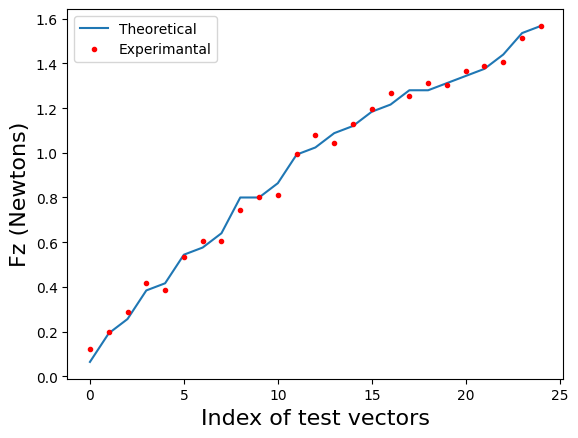}
        \caption{Force along z-axis}
        \label{fig:Fz}
    \end{subfigure}
    % \hfill % Add some space between the subfigures
    \begin{subfigure}{0.31\textwidth} % Adjust the width as needed
        \centering
        \includegraphics[height=4cm]{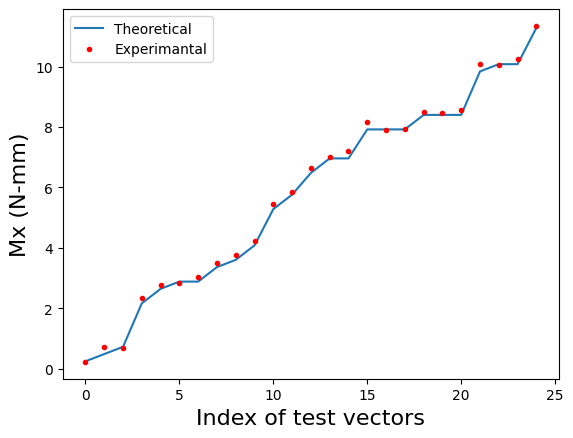} % Replace "your_second_image.eps" with the filename of your second image
        \caption{Moment along x-axis}
        \label{fig:Mx}
    \end{subfigure}
    % \hfill % Add some space between the subfigures
    \begin{subfigure}{0.31\textwidth} % Adjust the width as needed
        \centering
        \includegraphics[height=4cm]{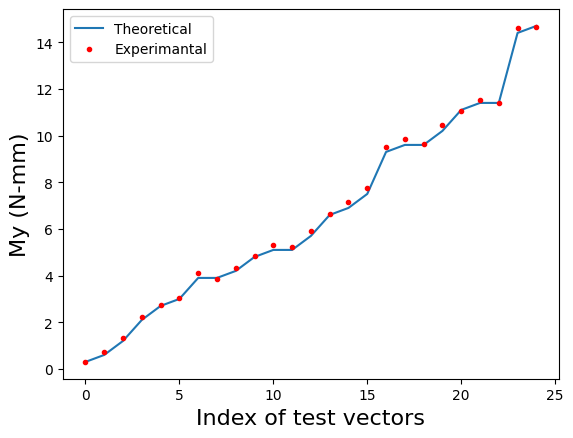} % Replace "your_second_image.eps" with the filename of your second image
        \caption{Moment along y-axis}
        \label{fig:My}
    \end{subfigure}
    \caption{Theoretical and Experimental results from testing.}
    \label{fig:experimental_results}
\end{figure}

The developed three-axis force/torque sensor was tested by
applying different values of forces and moments. The results
are shown in Fig. \ref{fig:experimental_results}. The accuracy of the
developed force/torque sensor was found out to be around
95\%. It was observed that the experimental values for $F_z$ were slightly above and below the theoretical values, while the experimental values for $M_x$ and $M_y$ were generally marginally higher than the theoretical values.

\section{Tactile Feedback}

Humans can perceive both frequency and intensity changes
in a vibration signal. Consequently, a variation in force can be
communicated to the surgeon by changing the frequency or
intensity of a vibration signal \cite{13_schoonmaker2006vibrotactile}. A tactile feedback system
outputting different intensities of vibration for different ranges
of gripping force between surgical tool tip and tissue was
designed to provide surgeons with better haptic feedback.
   
\begin{figure} [h]
   \begin{center}
   \begin{tabular}{c} %% tabular useful for creating an array of images 
   \includegraphics[height=5cm]{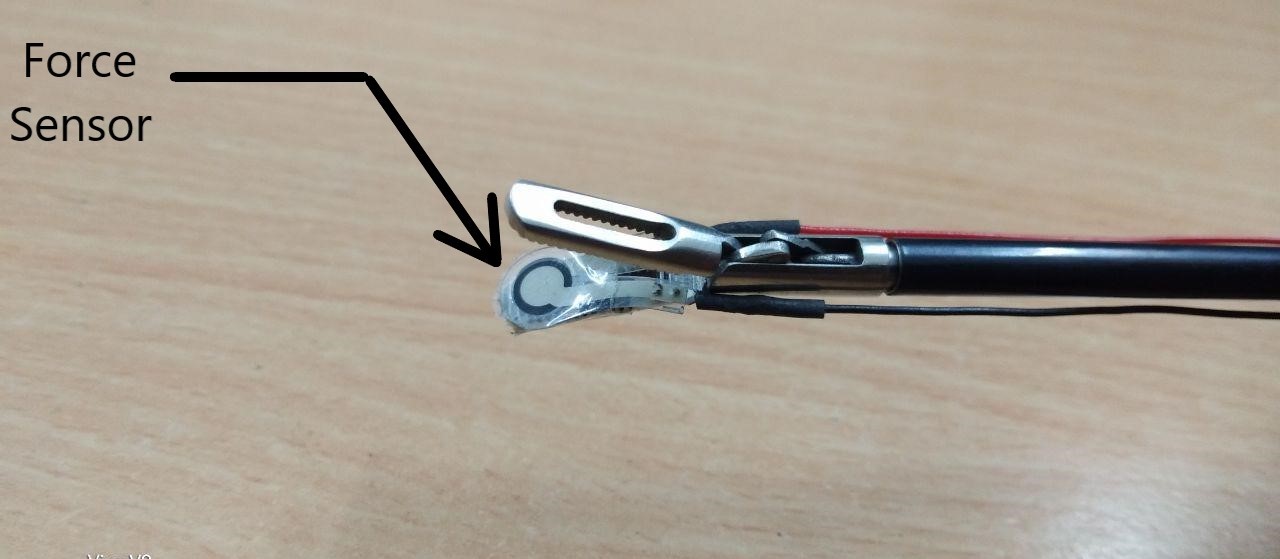}
   \end{tabular}
   \end{center}
   \caption[example] 
   { \label{fig:Flexiforce} 
Force sensor integrated to surgical tool tip }
   \end{figure}

A commercially available force sensor, FlexiForce™ A101
Sensor (TekScan, Inc., South Boston, Massachusetts, USA)
was used for estimating the gripping force. The sensing area of
sensor is a circle with 5mm diameter. The sensor is attached to
the inner side of the surgical tool tip as shown in Fig. \ref{fig:Flexiforce}. This setup is intended to aid in the development and refinement of surgical skills, not for use on patients.

\begin{figure} [h]
   \begin{center}
   \begin{tabular}{c} %% tabular useful for creating an array of images 
   \includegraphics[height=5cm]{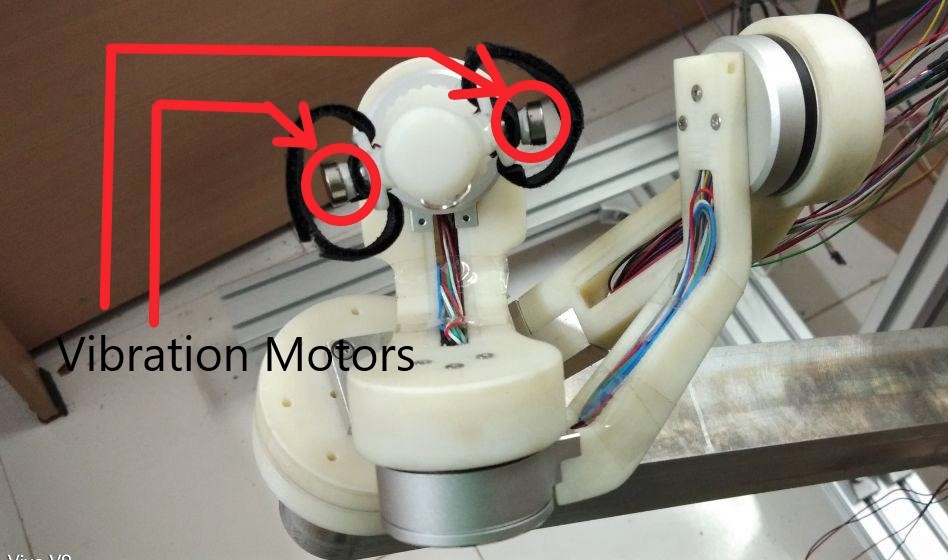}
   \end{tabular}
   \end{center}
   \caption[example] 
   { \label{fig:ERMmaster} 
Vibration motors integrated to finger clutch on master arm }
   \end{figure}

   The value of the gripping force is obtained from the force
sensor and is further used for providing vibration feedback to
the surgeon. Two commercially available eccentric rotating
mass vibration motors are integrated to the finger clutch of
each manipulator of the master side to provide vibration
feedback. The arrangement is shown in Fig. \ref{fig:ERMmaster}.

In the designed tactile feedback system, each manipulator on the master side has two vibration motors attached to the finger clutch. One vibration motor is programmed to provide a linear vibration response proportional to the force sensed by the force sensor, giving the surgeon feedback on changes in the applied gripping force between the surgical tool tip and the tissues. The second vibration motor activates at the highest vibration frequency only when the sensed force exceeds a threshold safe limit, which the surgeon can set before the procedure. This ensures that the surgeon does not apply excessive force, thereby preventing tissue damage.

\section{Conclusion}

In this study, a master-slave robotic endotrainer, akin to the DaVinci surgical robot, was developed to provide low-cost robotic surgical training for surgeons. Both kinesthetic and tactile feedback systems were effectively implemented in the robotic endotrainer to enhance the training experience.

A novel, low-cost three-axis force torque sensor was designed and integrated seamlessly into the surgical tool on the slave side. This sensor had 95\% accuracy in capturing force information from the interaction between the surgical tool and its environment. The acquired force data was utilized to provide kinesthetic feedback, allowing surgeons to gain a better understanding of the forces at play during surgical procedures. The developed force/torque sensor not only enhances the capabilities of the robotic endotrainer but also holds potential for broader applications. With slight modifications to its mechanical design and ruggedness, this sensor could be adapted for use in industrial robots, opening avenues for its deployment in various fields requiring precise force measurement and feedback.

Additionally, a simple yet effective vibration-based tactile feedback system was developed. This system offered clear indications of the gripping forces between the surgical tool tip and the tissues, ensuring that the user could discern subtle changes in force application, thereby improving their tactile sensitivity and precision.

Overall, this study successfully demonstrates the feasibility of developing a cost-effective robotic endotrainer that integrates advanced haptic feedback systems, contributing to the improvement of surgical training.

% In this study, a master-slave robotic endotrainer similar to
% the DaVinci surgical robot was developed with the aim of
% providing low cost robotic surgical training to surgeons.
% Kinesthetic and Tactile feedback systems were implemented
% effectively on the robotic endotrainer.

% A novel low cost three-axis force torque sensor was
% developed. The sensor was easily integrated into the surgical
% tool on the slave side. The sensor was able to obtain force
% information from the interaction between the surgical tool and
% the and its environment with 95\% accuracy. This force
% information was used to provide kinesthetic feedback. A
% simple and effective vibration based tactile feedback system
% was also developed that provided clear indication of the
% gripping forces between the surgical tool tip and the tissues.
% The developed force/torque sensor can be used for other
% applications such as in industrial robots by a slight
% modification of the mechanical design.

\acknowledgments % equivalent to \section*{ACKNOWLEDGMENTS}       
 
The authors acknowledge the DST-GITA funded, India -
Republic of Korea Joint applied R\&D Programme titled
“Design and Development of Robotic Endotrainer” (Ref.No:
GITA/DST/IKCP201401-005/2015, DT: 07.05.2015) for
financial support for this work.

% References
\bibliography{report} % bibliography data in report.bib
\bibliographystyle{spiebib} % makes bibtex use spiebib.bst

\end{document}